%% file: main.tex
\documentclass{article}

\input{macros/common}
\input{macros/names}
\input{macros/comments}

\usepackage[accepted]{mlsys2024}

\mlsystitlerunning{{Asynchronous LLM Function Calling}}

\begin{document}

\twocolumn[
\mlsystitle{Asynchronous LLM Function Calling}



\mlsyssetsymbol{equal}{*}

\begin{mlsysauthorlist}
\mlsysauthor{In Gim}{yale}
\mlsysauthor{Seung-seob Lee}{yale}
\mlsysauthor{Lin Zhong}{yale}
\end{mlsysauthorlist}

\mlsysaffiliation{yale}{Department of Computer Science, Yale University, New Haven, United States}

\mlsyscorrespondingauthor{In Gim}{in.gim@yale.edu}

\mlsyskeywords{Machine Learning, MLSys}

\vskip 0.3in

\begin{abstract}
Large language models (LLMs) use function calls to interface with external tools and data source. However, the current approach to LLM function calling is inherently synchronous, where each call blocks LLM inference, limiting LLM operation and concurrent function execution.
In this work, we propose \sys, a system for asynchronous LLM function calling. \sys improves LLM's operational efficiency by enabling LLMs to generate and execute function calls concurrently. Instead of waiting for each call's completion, \sys introduces an interrupt mechanism to asynchronously notify the LLM in-flight when function calls return. 
We design an in-context protocol for function calls and interrupts, provide fine-tuning strategy to adapt LLMs to the interrupt semantics, and implement these mechanisms efficiently on LLM inference process.
We demonstrate that \sys can reduce end-to-end task completion latency from 1.6\x-5.4\x compared to synchronous function calling on a set of benchmark tasks in the Berkeley function calling leaderboard (BFCL). Furthermore, we discuss how interrupt mechanisms can be extended to enable novel human-LLM or LLM-LLM interactions.
\end{abstract}
]

\pagestyle{plain}


\printAffiliationsAndNotice{} 

\input{sections/introduction}
\input{sections/background}

\input{sections/methodology}

\input{sections/implementation}
\input{sections/evaluation}
\input{sections/discussion}
\input{sections/conclusion}
\newpage
\bibliography{abr-short,main}
\bibliographystyle{mlsys2024}
\clearpage
\appendix
\input{sections/appendix}

\end{document}

%% file: macros/common.tex
\usepackage{microtype}
\usepackage{graphicx}
\usepackage{booktabs} 
\usepackage{amsfonts}
\usepackage{amsmath}
\usepackage{listings}
\usepackage{xcolor}
\usepackage{float}
\usepackage{tabularx} 
\usepackage{caption}
\usepackage{subcaption}
\usepackage{adjustbox}
\usepackage{enumitem}
\usepackage{tikz}

\usepackage{hyperref}
\usepackage{array}
\usepackage{colortbl}
\usepackage{xurl}
\usepackage{color,soul}
\usepackage{xspace}
\usepackage{comment}

\definecolor{pblue}{RGB}{78,121,167}
\definecolor{pred}{RGB}{225,87,89}
\definecolor{pgreen}{RGB}{89,161,79}
\definecolor{darkgreen}{rgb}{0,0.5,0}

\newcommand*\bcircled[1]{\tikz[baseline=(char.base)]{
    \node[shape=circle,fill,inner sep=0.5pt] (char) {\textcolor{white}{\textit{#1}}};}}

\newcommand{\paragraphb}[1]{\vspace{0.05in}\noindent{\bf #1.}}

\lstdefinestyle{xmlstyle}{
  basicstyle=\ttfamily\scriptsize,
  morestring=[s]{"}{"},
  morecomment=[s]{?}{?},
  morecomment=[s]{!--}{--},
  commentstyle=\color{darkgreen},
  moredelim=[s][\color{black}]{>}{<},
  moredelim=[s][\color{pred}]{\ }{=},
  stringstyle=\color{pgreen},
  identifierstyle=\color{pblue},
  tabsize=2,
  breaklines=true,
  frame=single,
  xleftmargin=3.4pt,
  xrightmargin=3.4pt
}
\newtheorem{theorem}{Theorem}[section]

\newcommand{\code}[1]{{\texttt{#1}}}

%% file: macros/names.tex
\newcommand{\sys}{AsyncLM\xspace}
\newcommand{\dsl}{CML\xspace}
\newcommand{\sync}{\emph{Sync}\xspace}
\newcommand{\synccom}{\emph{Sync-Parallel}\xspace}
\newcommand{\asyncn}{\emph{Async-Naive}\xspace}
\newcommand{\async}{\emph{Async}\xspace}
\newcommand{\x}{$\times$\xspace}

%% file: macros/comments.tex
\definecolor{lightgray}{gray}{0.9}
\definecolor{lightblue}{rgb}{0.9,0.9,1}
\definecolor{blue_bg}{rgb}{0.7,0.85,1}
\definecolor{lightyellow}{rgb}{1,1,0.8}
\definecolor{lightpurple}{rgb}{1,0.85,1}
\definecolor{lightgreen}{rgb}{0.8,1,0.7}
\definecolor{red}{rgb}{1,0,0}
\definecolor{darkgreen}{rgb}{0.4,0.7,0.3}
\definecolor{darkblue}{rgb}{0.2,0.7,0.9}


%% file: sections/introduction.tex
\section{Introduction}
\label{sec:introduction}


Function-calling capabilities enable large language models (LLMs) to access external data sources and tools, such as weather forecasts and calculators. Both commercial and open-source LLMs have integrated this feature~\cite{schick2024toolformer,patil2023gorilla}, unlocking new possibilities for diverse applications, from autonomous AI agents operating in dynamic environments~\cite{wang2024survey} to neurosymbolic systems combining symbolic reasoning with LLMs to solve complex problems~\cite{trinh2024solving}.

LLM function calls are synchronous, with the LLM and the function call executor taking turns generating and executing calls. Although simple to implement, this approach is neither resource-efficient nor responsive. Each function call blocks LLM inference---one of the most resource-intensive processes---until the function returns. 
From the executor's perspective, this limits concurrency since all function calls must finish in the order they are initiated by the LLM.
These inefficiencies worsen as the number of functions increases with the complexity of the task~\cite{compound-ai-blog}.


\begin{figure}[t]
    \centering
    \includegraphics[width=0.95\columnwidth]{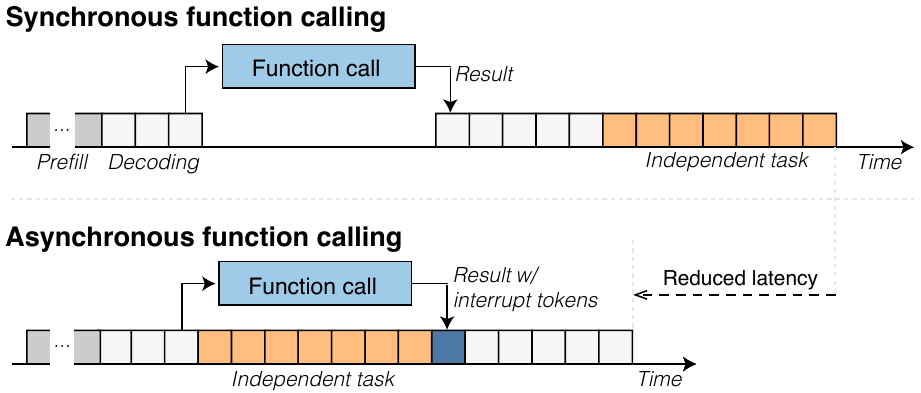}
    \vspace{-0.5em}
    \caption{\textbf{Synchronous vs. asynchronous function calling.} Asynchronous function calling improves LLM's operational efficiency, by enabling the LLM to continue generating tokens for independent tasks while the function call executes in the background.}
    \label{fig:hook}
\end{figure}

Several studies have tried to address these challenges, including using compilers to parallelize function calls~\cite{kim2023llm}, fusing sequential calls to reduce overhead~\cite{singh2024llm}, designing concise call syntax~\cite{chen2023typefly}, and optimizing LLM serving systems for function calling~\cite{abhyankar2024infercept,gao2024fast,xu2024conveyor}. While these methods help reduce function execution time or the number of function calls (\S\ref{sec:background}), they are fundamentally limited by the synchronous nature of function call, e.g., the LLM waiting for the function call executor to finish.

We propose \sys, a system that enables \emph{asynchronous} interactions between LLMs and function call executors to overcome these limitations.
In \sys, the LLM and the executor operate independently without blocking each other, drawing inspiration from asynchronous programming paradigms where events, e.g., function call completions, occur independently of the main program flow, e.g., LLM token generation stream. \autoref{fig:hook} illustrates the concept.

The key mechanism of \sys is \emph{interruptible} LLM decoding. In \sys, function calls can be non-blocking; when a function call returns, the executor asynchronously notifies the LLM by injecting \emph{interrupt tokens} into the LLM's token generation stream. Although the concept is simple, \sys must address two challenges in coordinating the LLM and the executor.
(\textit{i}) The interrupt token must be carefully timed to avoid interfering with other tasks or function calls the LLM is generating. For example, injecting an interrupt halfway through an argument for another function call could disrupt both the ongoing function call generation and the returned function call.
(\textit{ii}) The LLM must correctly handle the injected interrupt, even though such interrupts do not follow the conversational patterns on which the model was trained.

To address these challenges, we co-design the interface for asynchronous function calls and the LLM fine-tuning strategy. (\textit{i}) We develop a token-efficient domain-specific language called \dsl to represent function calls and interrupts (\S\ref{sec:dsl}). \sys detects \dsl syntax during token generation to identify the start and end of function calls, deferring interrupts accordingly. \dsl also helps the LLM resume processing after an interrupt by embedding the necessary context within the interrupt syntax.
(\textit{ii}) We overcome the second challenge by fine-tuning the LLM to teach it how to generate asynchronous function calls, how to resume from handling an interrupt, and how to notify the LLM serving system when it must pause and wait for function call returns (\S\ref{sec:training}). All three cases use \dsl as the interface. Additionally, when the LLM chooses to wait, \sys determines the most efficient strategy to manage the blocking state, such as dropping, swapping, or retaining the KV cache.

The advantages of \sys over synchronous function calling are twofold. First, \sys reduces the execution time by overlapping the generation and execution of functions. This property ensures that \sys is at least as fast as synchronous function calling with parallel execution, even in the worst case (\S\ref{ssec:math}). Second, \sys enables automatic parallelism without requiring prior knowledge of the dependency graph for future function calls. For example, given the task, ``Tell me if Leonhard Euler is one of my academic ancestors using the math genealogy API,'' and an API definition, \texttt{find\_advisors(str)$\rightarrow$list[str]}, \sys can recursively and in parallel invoke \texttt{find\_advisors}, similar to performing a parallel depth-first search.
In contrast, efficiently handling such tasks in a synchronous function calling scheme is challenging, as it requires determining the list of parallelizable function calls in advance (\S\ref{sec:background}).

We implement \sys on Llama 3 models~\cite{dubey2024llama} and emulate fine-tuning on GPT-4o using in-context prompting—where the model is guided through \dsl examples and instructions in the input prompt (\S\ref{sec:implementation}). We evaluate both models on a suite of function calling tasks from the Berkeley Function Calling Leaderboard (BFCL)~\cite{bfcl} (\S\ref{sec:evaluation}). Benchmark comparisons show that \sys accelerates end-to-end task completion by up to 1.6\x-5.4\x over synchronous sequential function calling and achieves a speedup of up to 2.1\x over synchronous parallel function calling. We also assess the impact of \sys on function calling accuracy, demonstrating that \sys maintains the same level of accuracy as synchronous function calling in fine-tuned Llama models. Notably, GPT-4o can handle asynchronous function calling without explicit fine-tuning. Finally, we discuss the potential of the introduced interrupt mechanism for use in novel AI applications involving human-LLM or LLM-LLM interactions, such as interruptible LLM assistants.

\begin{figure*}[t]
    \centering
    \includegraphics[width=2\columnwidth]{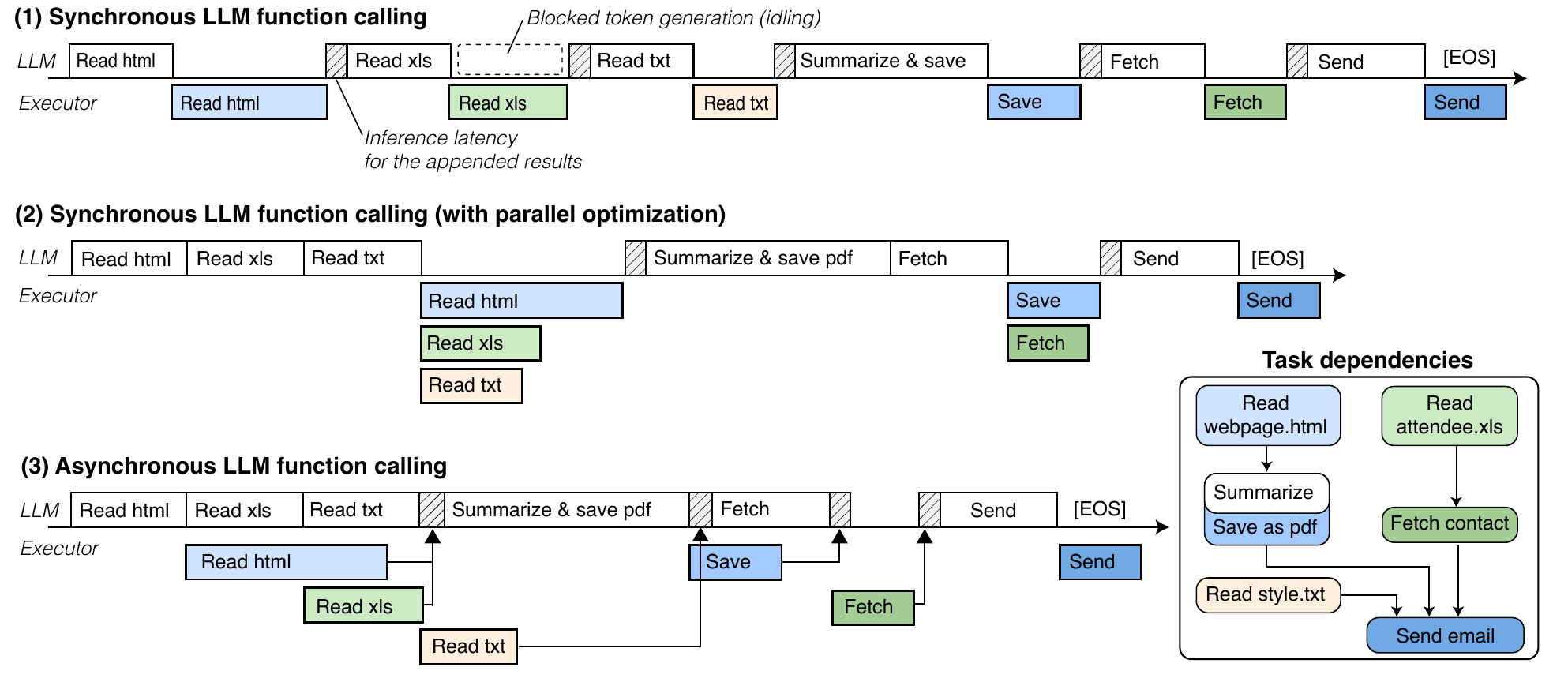}
    \caption{\textbf{Comparison of LLM-executor interactions.} Parallel function calling reduces end-to-end execution time by generating function calls upfront and executing them in parallel, while still requiring the LLM to wait for all calls to complete. Asynchronous function calling overcomes this limitation by allowing the LLM to generate tokens while function calls execute in the background. \textbf{The task:} \emph{``Summarize webpage.html into event.pdf, email it (styled per style.txt) to people in attendee.xls, and CC the dept chair from the directory if not listed.''}}
    \label{fig:related_work_comparison}
\end{figure*}

%% file: sections/background.tex
\section{Background and Related Work}
\label{sec:background}\label{sec:overview}

The concept of augmenting LLMs with code execution~\cite{mialon2023augmented} has been extensively explored, notably in contexts like retrieval-augmented generation~\cite{khattab2022demonstrate,yao2024react}, autonomous agents~\cite{huang2024understanding,wang2024survey}, and neurosymbolic problem solving~\cite{pan2023logiclm,trinh2024solving}. 

\paragraphb{Learning to generate function calls}
The most common method for LLM interaction with external systems is tool or function calling~\cite{schick2024toolformer,shen2024hugginggpt}, where the model autonomously generates calls to external executors (e.g., API servers, code interpreters). This ability is refined either through fine-tuning on diverse task datasets~\cite{patil2023gorilla} or using in-context instructions~\cite{liang2024taskmatrix}. Function descriptions, including arguments, are provided in a structured prompt format, often in JSON. Our approach builds on this paradigm by enabling asynchronous function calling, requiring LLMs to (1) consider execution time in call generation and (2) use interrupt semantics to decide subsequent calls.

\paragraphb{Efficient LLM function calling}
A major challenge in function calling is optimizing efficiency to enhance resource utilization and reduce latency. Various studies have explored different optimization strategies to improve the function calling process. For instance, parallel function calling approaches~\cite{kim2023llm,openai_function_calling} instruct the LLM to bundle calls that can be executed simultaneously, enabling external compilers to optimize these batches for parallel execution. Sequential function call optimizations include methods like function call fusion~\cite{singh2024llm}, caching~\cite{singh2024an}, compact syntax for call representation~\cite{chen2023typefly}, and partial execution of function calls~\cite{xu2024conveyor}, which allow overlapping of generation and execution of a single code block.

\paragraphb{Limitations of a synchronous interaction}
Currently, LLMs perform synchronous function calls. This interleaved nature of generation and execution introduces extra overheads in LLM inference, due to the stateless nature of LLMs, which require a new session per function call. This has motivated research on efficient management of KV cache usage during waiting periods~\cite{abhyankar2024infercept,yu2023stateful,gao2024fast}. More importantly, synchronous function calling fundamentally limits resource utilization, since either only token generation or function execution can happen at the same time. ReWOO~\cite{xu2023rewoo} addresses this limitation by prompting LLMs to decouple reasoning from observation to consolidate the execution of multiple function calls into one. Our work is conceptually aligned with ReWOO but remains agnostic to the LLM's reasoning strategy, making it more broadly applicable.

\paragraphb{Our approach}
\sys addresses the limitations of synchronous function calling by enabling asynchronous execution. \autoref{fig:related_work_comparison} illustrates the benefits of asynchronous function calling over synchronous schemes. While function parallelization optimizations~\cite{kim2023llm,openai_function_calling} reduce end-to-end execution time, the LLM must still wait for function calls to complete, leaving resources idle. Asynchronous function calling eliminates this bottleneck by allowing the LLM to continue generating tokens for independent tasks alongside ongoing function calls. To support asynchronous calls, \sys introduces mechanisms to asynchronously notify LLMs in-flight when funtion call returns, and fine-tunes the LLM to understand interrupts and leverage concurrent execution.

\sys is built around two key innovations. First, it uses a domain-specific language (\dsl) to describe asynchronous function calls and interrupts, embedding the necessary context for seamless integration (\S\ref{sec:dsl}). Second, its fine-tuning strategy trains the LLM to respond to asynchronous events, generate concurrent function calls, and notify the serving system when it needs to pause and wait for previous calls to complete (\S\ref{sec:training}) to ensure correct function call dependencies. We implement these components in \sys’s LLM serving system, which monitors token generation, schedules function calls on the executor, and manages the token stream by injecting interrupts or pausing generation as needed (\S\ref{sec:implementation}).

%% file: sections/methodology.tex
\section{Representing Asynchronous Interaction with \dsl}
\label{sec:dsl}

\begin{figure*}[!t]
    \centering
    \includegraphics[width=1.8\columnwidth]{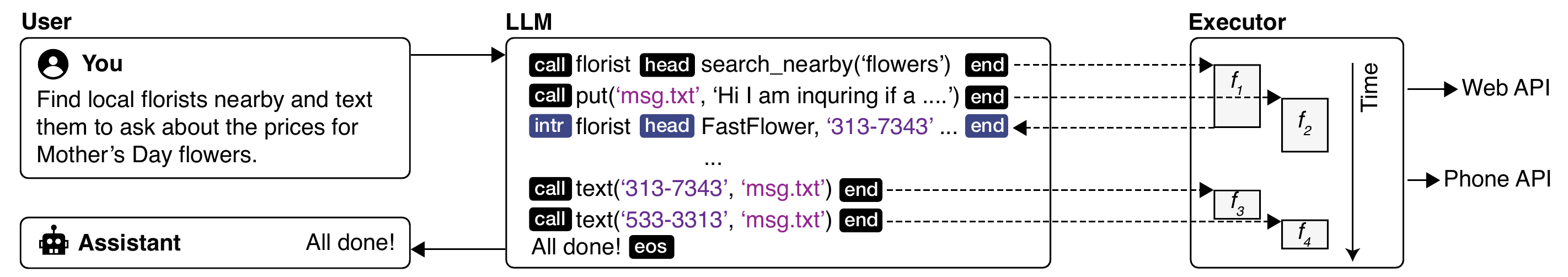}
    \vspace{-0.5em}
    \caption{\textbf{Asynchronous function execution workflow.} The LLM generates function calls in \dsl format, which are monitored real-time and sent to the code executor in the background. When a function call completes, the executor asynchronously notifies the LLM by inserting an interrupt along with the execution result. 
    } 
    \label{fig:overview}
\end{figure*}

We define a simple domain-specific language, called \emph{Context Markup Language} (\dsl), to represent asynchronous function calls and interrupts. \dsl acts as the interface between the LLM and the executor, with its syntax ensuring that each component provides the necessary context when interacting through this interface. For example, the LLM can generate a function call in \dsl to notify the executor, while the executor signals completion by inserting interrupt tokens in \dsl, as demonstrated in \autoref{fig:overview}.

\dsl uses a minimal set of specialized tokens: \texttt{[CALL]}, \texttt{[INTR]}, \texttt{[TRAP]}, \texttt{[END]}, and \texttt{[HEAD]}. The \texttt{[CALL]}, \texttt{[INTR]}, and \texttt{[TRAP]} tokens initiate a \emph{control block}, which represents a function call (\S\ref{ssec:async_ftncall}), an interrupt (\S\ref{ssec:async_trap}), or a trap (a special interrupt), respectively. The \texttt{[END]} token marks the end of the control block, and \texttt{[HEAD]} separates optionally provided metadata, such as a function call identifier, and the body of function call. The rest of this section defines the semantics of \dsl.

\subsection{Initiating Function Calls}\label{ssec:async_ftncall}

In \sys, the LLM initiates function calls using the following format: \texttt{[CALL] {function call} [END]}. The function call can be written in any valid executable language, such as Python code or a JSON abstract syntax tree (AST), supported by the executor. However, only one language should be used consistently within each function call block. Independent function calls should be placed in separate blocks to allow for parallel execution.

The generated function calls initiate execution without blocking the token stream, enabling implicit parallelism. For example, as shown in \autoref{fig:overview}, if the LLM generates two function calls (\texttt{search\_nearby} and \texttt{put}), the executor processes each in a separate worker, allowing pipelined generation and execution of function calls. This overlap reduces latency, as the execution of \texttt{search\_nearby} can occur simultaneously with the generation of writing text messages. Function calls are generated according to their dependency order, ensuring that dependent calls are executed in the correct sequence. For instance, the LLM can generate a function call, \texttt{text}, only after searching for florists and composing the message are done.

\paragraphb{Assigning identifiers for interrupts}
If the LLM will need to refer to the function result for subsequent calls or reasoning, it can include an identifier in the function header (e.g., \texttt{[CALL] job1 [HEAD] {function call} [END]}), where \texttt{job1} acts as a unique identifier. To avoid conflicts, the LLM generates identifiers following Python variable naming conventions. Identifiers must remain unique throughout the session. Although our prototype does not implement this, a uniqueness check could be included as part of syntax validation (\S\ref{ssec:token}).




\subsection{Trigerring Interrupts}
\label{ssec:async_interrupt}\label{ssec:async_trap}

When the executor completes a function call with a registered identifier, it asynchronously notifies the LLM by inserting an interrupt block at the end of the token stream (i.e., the LLM context). Token generation resumes after the interrupt block is added. The format of an interrupt block is: \texttt{[INTR] {id} [HEAD] {value} [END]}, where \texttt{{id}} matches the identifier from the corresponding \texttt{[CALL]} block (e.g., \texttt{job1} from the earlier example). The \texttt{{value}} contains the executor's result, such as the function's output or an error message.

\paragraphb{Critical sections}
Since interrupts modify the LLM’s token generation flow, it is essential to ensure that the context following an interrupt conforms to \dsl syntax. For example, if an interrupt block is inserted during the generation of a function call, the resulting tokens could become logically incorrect, violating \dsl syntax. To prevent this, \sys temporarily disables interrupts during specific token generation periods, inspired by how operating systems defer lower-priority interrupts while handling others. A flag called \emph{critical section}, implemented within the interrupt manager (\S\ref{ssec:intr}), determines whether the executor can insert interrupts. This flag is set to \texttt{false} while a function call block is being generated and resets to \texttt{true} once the LLM exits the block. Any interrupts that occur during a critical section are queued and inserted when the flag is set to \texttt{true}.


\paragraphb{Waiting for interrupts}
In synchronous function calling, the LLM stops generating tokens after making a function call, waiting for the call to complete. In contrast, asynchronous function calling lets the LLM continue generating tokens for other tasks without waiting. However, the LLM must sometimes pause to wait for function results when inter-task dependencies exist. The standard \texttt{[EOS]} token is not sufficient for this purpose because it cannot indicate whether token generation is complete and resources can be released, or if the LLM is temporarily pausing until the function result arrives.

To resolve this ambiguity, \sys introduces self-initiated interrupts, called \emph{traps}, which notify the LLM serving system (\S\ref{ssec:trap_handler}) when to pause token generation. Each trap follows the simple structure \texttt{[TRAP][END]}. Traps create explicit boundaries for asynchronous function calls, helping generate training samples to fine-tune LLMs for asynchronous function handling (\S\ref{sec:training}). They also enable optimization opportunities for the serving system (\S\ref{sec:implementation}).


\section{Learning to Handle Interrupts}
\label{sec:training}


We propose a fine-tuning scheme to train LLMs to (\textit{i}) generate asynchronous function calls and traps using \dsl and (\textit{ii}) handle interrupts that deliver the results of previous function calls. The core idea is to construct a dataset with simulated function calls and interrupts that model ideal interactions between the LLM and the executor. We extract task descriptions and function definitions from publicly available function-calling datasets, add estimated completion times, and present them to the LLM in JSON format as part of the prompt~\cite{openai_function_calling,patil2023gorilla,liang2024taskmatrix}. For simplicity, we assume that each task can be completed with a finite number of function calls.

\subsection{Training objectives}
The primary objective of fine-tuning in \sys is to train the LLM to minimize the total task completion time (i.e., the makespan) by effectively utilizing asynchronous function calling, while respecting task dependencies and considering estimated function execution times. Specifically, the LLM must make decisions in the following areas.

\paragraphb{Deciding the next function call} 
The LLM selects the next function to call based on the context.\footnote{We use the \emph{context} to refer to the sequence of tokens currently visible to the LLM.} It identifies which functions are available to be called immediately---those without any pending dependencies---and chooses the most suitable one. \sys uses a Longest-Processing-Time-first (LPT) strategy, where the LLM prioritizes calling the function with the longest estimated execution time among those that are ready. This approach reduces idle time by maximizing the overlap between function call generation and execution, making it particularly effective when multiple independent functions can be called. LPT is optimal for scenarios with parallel function calling (\S\ref{sec:evaluation}).

\paragraphb{Handling interrupts}
When an interrupt appears in the context, the LLM must decide whether to continue generating tokens for the current task or shift to addressing the interrupted task. For example, the LLM might ignore the interrupt if the completed task has lower priority than the current task or if it was the final task in a sequence. \sys handles interrupts using the same priority principles as for deciding the next function call. An interrupt may introduce new functions that are ready to be called; if a newly available function has the longest estimated processing time among the options, the LLM is trained to call it next. For instance, in the task scheduling scenario shown in \autoref{fig:related_work_comparison}, when ``Read html'' and ``Read xls’’ complete, the LLM chooses to do ``Summarize \& save pdf'' first over ``Fetch contact,’’ prioritizing the function with the longer estimated processing time.

\paragraph{Generating Traps.}  
When no functions are available to call due to dependencies on future interrupts, and there are no other tokens to generate, the LLM must generate a trap to pause token generation.

\subsection{Generating Training Samples}
To create training samples, we simulate ideal interactions between the LLM and executor using existing function-calling traces from LLM benchmarks with multi-turn scenarios \cite{lu2024toolsandbox,bfcl,yao2024tau}.

\paragraphb{DAG generation}
We developed a Python program to extract directed acyclic graphs (DAGs) of function calls from each sample in the dataset. In sequential scenarios, the DAG is linear, with nodes representing function calls and edges indicating dependencies. In parallel scenarios, the DAG branches to represent independent function calls. Multi-turn scenarios consist of multiple DAGs.

\paragraphb{Simulating interrupts}
We simulate interactions between the LLM and the executor on these DAGs using the LPT strategy. In each simulation run, we assign random estimated execution times to function calls, ranging from 1 ms to 1 s, to prevent the model from overfitting to specific function names. These estimates are provided to the LLM as part of the input prompt. To determine interrupt timing, we randomly set a time-per-output-token (TPOT) between 5 ms and 30 ms per simulation and track elapsed time by counting generated tokens. When a function call completes, we insert an interrupt block containing the execution result. If all functions in the DAG are waiting on dependencies, we insert a trap block and inject the next interrupt.

%% file: sections/implementation.tex
\begin{figure*}[!ht]
    \centering
    \includegraphics[width=2\columnwidth]{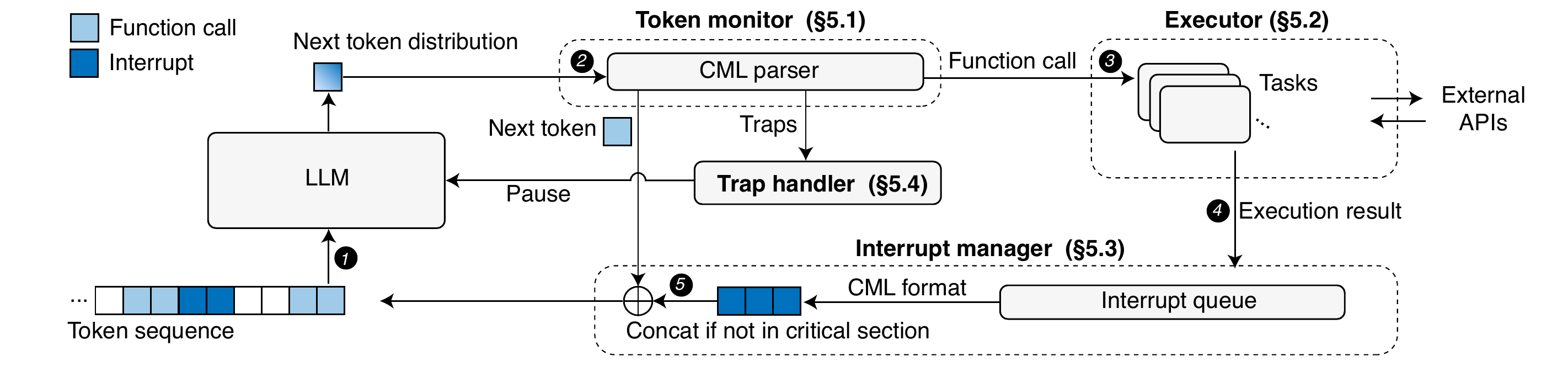}
    \vspace{-0.7em}
    \caption{\textbf{Overview of \sys's LLM inference process.} LLMs generate tokens autoregressively. \sys implements asynchronous function calling by augmenting this process with additional steps to monitor the token generation process for \dsl recognition and manage asynchronous function calls, interrupts, and traps based on recognized \dsl syntax. The numbered steps, \bcircled{1} to \bcircled{5}, show the logical flow from generating a function call to executing it and returning the result.}
    \label{fig:system}
\end{figure*}

\section{Implementation}
\label{sec:implementation}
We implement \sys by intercepting the LLM’s autoregressive token generation, as illustrated in \autoref{fig:system}. The system is built using the Python {transformers} library~\cite{wolf2020transformers}. \sys consists of four main components: a token monitor, an executor, an interrupt manager, and a trap handler. In this section, we describe the implementation of each component and the modifications made to the LLM inference. While some aspects of these components can be implemented on cloud-based LLM APIs without modifying the serving system (\S\ref{ssec:impl_cloud_api}), such implementations are practical only in certain scenarios (\S\ref{ssec:multistep}).

\subsection{Token Monitor}
\label{ssec:token}

The token monitor audits and regulates the LLM’s token generation process. Its primary functions are (\textit{i}) to notify the executor or trap handler immediately when a function call or trap is detected in the token stream, and (\textit{ii}) to enforce \dsl syntax compliance.
Implemented as a hook into the LLM's next-token sampling process, the token monitor constrains the next-token distribution to ensure generated tokens follow valid \dsl syntax~\cite{geng2023grammar}. It uses a finite-state machine (FSM) to enforce \dsl rules and blocks the LLM from producing \texttt{[INTR]} tokens, thereby preventing ``hallucinated'' interrupts, as interrupts must be injected by the system rather than generated by the LLM. After sampling each token, the token monitor checks for a function call or trap and, if detected, immediately notifies the executor (\S\ref{ssec:executor}) or trap handler (\S\ref{ssec:trap_handler}). Additionally, after determining whether the LLM is in an interruptible state, the token monitor sets a critical section flag accordingly. It sends both the generated token and the critical section flag to the interrupt manager (\S\ref{ssec:intr}).

\subsection{Executor}
\label{ssec:executor}

The executor manages the execution of function calls generated by the LLM. It receives each function call from the token monitor (formatted in Python syntax in our prototype) along with an optional identifier used to manage future interrupts. Each function call runs on a dedicated worker, allowing multiple calls to execute concurrently if resources permit. These workers interact directly with external systems, such as API servers or code interpreters. Once a function completes, the executor sends the result and identifier to the interrupt manager (\S\ref{ssec:intr}).


\subsection{Interrupt Manager}
\label{ssec:intr}
The interrupt manager has three main functions: (\textit{i}) managing an interrupt queue, (\textit{ii}) tracking when the token generation process is interruptible based on a critical section flag, and (\textit{iii}) inserting \dsl-formatted interrupt blocks into the token stream. When the executor completes a function call, it adds the result to the interrupt queue. During each decoding step, the interrupt manager receives newly generated tokens and the critical section flag from the token monitor (\S\ref{ssec:token}). If the process is interruptible, i.e., the critical section flag is not set, it formats all queued interrupts in \dsl, tokenizes them, and appends them to the generated tokens. These tokens will then be processed by the LLM in the next decoding step.

\subsection{Trap Handler}\label{ssec:trap_handler}

The trap handler’s goal is to minimize idle KV cache usage in GPU memory without adding latency to task completion. When the token monitor detects a trap---indicating that token generation needs to be paused---it notifies the trap handler with the current context. Based on this notification, the trap handler determines the best strategy for managing the KV cache during the pause by considering the number of tokens in the current context and the estimated time until the next interrupt completes. Prior work has shown that recomputing the KV cache scales quadratically with the number of tokens, while swapping it to host memory scales linearly~\cite{abhyankar2024infercept,gim2024prompt}. Given the scaling characteristics of these costs, the trap handler keeps the KV cache in GPU memory if both recompute and swap times exceed the estimated wait time. Otherwise, it opts to recompute if recompute latency is lower, or to swap if swap latency is lower. We provide a real-world example of an ideal trap handling strategy in \S\ref{ssec:multistep}.

\subsection{Implementation on Chat Completion APIs}
\label{ssec:impl_cloud_api}

To demonstrate \sys’s adaptability, we also implement it using OpenAI’s (streaming) chat completion API without modifying the serving system. The executor implementation can be reused as is. For the token monitor, we implement only the \dsl parser without constrained decoding, as the API already streams sampled tokens. To emulate the interrupt insertion mechanism in the interrupt manager, we start a new request to the API server whenever an interrupt is triggered, discarding the previous session. The trap handler is unnecessary for cloud APIs, which are stateless and recompute the entire KV cache from scratch for each new session when an interrupt is inserted.\footnote{While some LLM services support caching prompts for reuse, these caches are typically designed for frequently accessed information, such as documents, rather than for runtime use.} We note that this implementation is practical only when the time-to-first-token (TTFT) latency is low, which is generally not the case for most cloud services (details in \S\ref{ssec:multistep}).

%% file: sections/evaluation.tex
\section{Evaluation}\label{sec:evaluation}
Our evaluation answers two questions: (\textit{i}) \emph{latency} (\S\ref{ssec:parallel}-- \S\ref{ssec:sys_overhead}), i,e., how much does asynchronous function calling reduce task completion latency compared to synchronous methods, and (\textit{ii}) \emph{correctness} (\S\ref{ssec:accuracy}), i.e., how does the asynchronous mechanism affect the correctness of generated function calls.

\textbf{Workloads.} 
To evaluate task completion latency and function calling accuracy, we use the Berkeley function calling leaderboard (BFCL)~\cite{bfcl}, which captures real-world function calls across eight domains, such as vehicle control, travel booking, file system operations, and the Twitter API. BFCL includes 84 unique functions.
We utilize three datasets from BFCL to cover different function calling scenarios: \code{v1-parallel}, \code{v2-parallel-live}, and \code{v3-base-multi-turn}. Specifically, \code{v1-parallel} and \code{v2-parallel-live} provide 400 parallel function calling scenarios, while \code{v3-base-multi-turn} offers 200 multi-step function calling scenarios.
To simulate more complex multi-step parallel function calling, we created a new dataset \code{v3-multi-step-parallel}. This dataset consists of 200 scenarios formed by randomly combining three distinct multi-step samples from the first round in \code{v3-base-multi-turn}. From the total of 800 samples, we use 200 for fine-tuning and the remaining 600 for evaluation. The measured execution time of each function ranges from 30 ms to 500 ms, with an average of 110 ms.

\textbf{\sys setup.}
We consider two LLM deployment settings: local and cloud. In the local deployment, LLM inference and function execution run on the same machine equipped with an NVIDIA RTX 4090 GPU. This deployment uses Llama-3.2 models~\cite{dubey2024llama} with 3B and 1B parameters. 
We fine-tune the Llama models following default configurations in LlamaFactory~\cite{zheng2024llamafactory}. Local models are served with Text-Generation-Interface~\cite{huggingface_text_generation_inference}.
In the cloud deployment, only function execution is local. This deployment uses OpenAI's GPT-4o and GPT-4o-mini.
 We adapt them using few-shot prompting; we select one example from each dataset and provide detailed instructions on their interpretation. 

\textbf{Baselines.} To compare against synchronous LLM function calling, we employ two synchronous baselines:\\
\indent $\bullet$ \sync: Sequential function calling where LLM code generation and execution are interleaved.\\
\indent $\bullet$ \synccom: Parallel function calling~\cite{openai_function_calling,kim2023llm}, where the LLM bundles independent function calls, and the executor runs them in parallel.\\
We also compare two implementations of \sys:\\
\indent $\bullet$ \asyncn: Based on the chat completion API without modifying the underlying LLM serving system.\\
\indent $\bullet$ \async: \sys with the co-designed inference process.

For \async latency measurements on cloud (GPT-4o), we report emulated results based on average token generation latency statistics (5 ms per output token) from the OpenAI API. This offers a rough estimation of the latency benefits of implementing \sys on cloud-based LLMs. 

Note that the officially reported accuracy of LLMs in generating function calls in BFCL is 62\% for GPT-4o and 43\% for Llama-3.2-3B~\cite{bfcl}. This can lead to biases in the latency evaluation, 
since incorrect function calls can miss some essential steps or have redundant steps.
To mitigate this interference in latency evaluations, we include a ``cheat sheet'' with ground truth answers in the prompt. This ensures consistency in the generated function calls across measurements.


\subsection{Parallel Function Calling}
\label{ssec:parallel}
\begin{figure}[t]
\centering
\includegraphics[width=0.9\columnwidth]{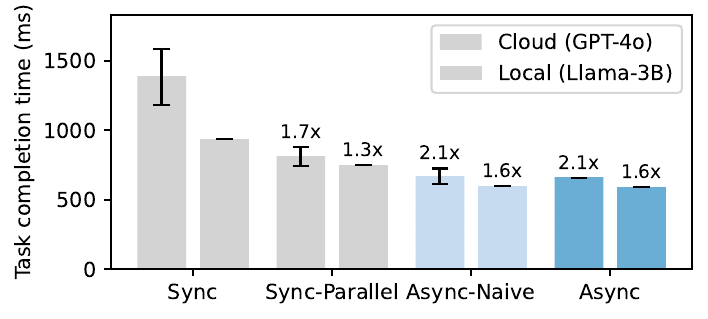}
\vspace{-1em}
\caption{\textbf{Parallel function calling latencies}. End-to-end task completion latency across the combined BFCL parallel dataset, with error bars representing the 10th and 90th percentile ranges. Bar annotations indicate speedups relative to \sync.}
\label{fig:parallel-latency}
\end{figure}

First we employ a simplistic setup where no function calls depend on 
each other and report results in \autoref{fig:parallel-latency}.
Parallel function calling is common~\cite{openai_function_calling}. For example, when a user asks ``\emph{Which city has a higher chance of rain tomorrow, Seattle or Vancouver?}'', the LLM can generate two function calls: one for weather data in Seattle and another for Vancouver, which can be executed in parallel.
We use \code{v1-parallel} and \code{v2-parallel-live} from BFCL for this evaluation. 

\textbf{Results.} \async improves latency by up to 2.1\x over \sync.
We measure the end-to-end task completion latency as the time between the first and last token generation. Our results show that \async completes tasks faster than \sync by 1.6\x for the local deployment and 2.1\x for cloud. In comparison, \synccom is 1.3\x faster than \sync for local and 1.7\x faster for cloud. Although \asyncn is slower than \async, it is still 1.2\x faster than \synccom for local. 

\begin{figure}[t]
    \centering
    \includegraphics[width=0.9\columnwidth]{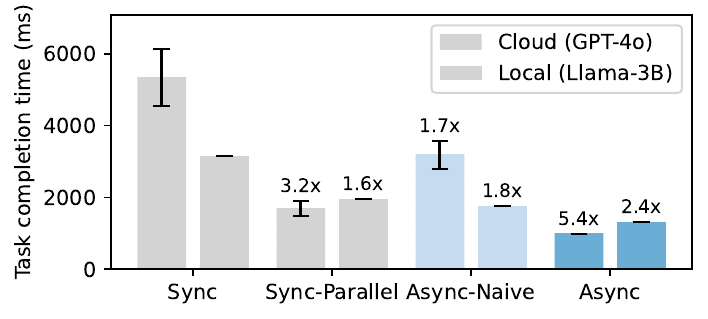}
    \vspace{-1em}
    \caption{\textbf{Multi-step parallel function calling latencies}. End-to-end task completion latency for simultaneously handling three parallel tasks in the BFCL multi-step dataset.}
    \label{fig:multitask-latency}
\end{figure}

\subsection{Multi-Step Parallel Function Calling}
\label{ssec:multistep}
For evaluations on more complex setup under function call dependencies, we evaluate \sys using the v3-multi-step-parallel dataset, which consists of three independent tasks requiring 1–5 sequential function calls each. The results are presented in \autoref{fig:multitask-latency}. In these function calling scenarios, \sys must respect task order dependencies and manage interrupt identifiers for each function call. For example, the task ``make pasta'' can involve two independent sequences: (\textit{i}) boil\_water() followed by put\_pasta\_noodles, and (\textit{ii}) chop\_vegetables() followed by stir\_fry(), and culminating in mix\_everything(). 

\textbf{Results.} \sys reduces latency by up to 5.4\x over \sync by parallelizing function calls in independent sequences; while \synccom reduces latency by 3.2\x compared to \sync. As illustrated in \autoref{fig:related_work_comparison}, unlike in \synccom, where the LLM needs to wait for all bundled functions to complete, \async uses the LPT strategy to schedule individual function calls. This enables the LLM to optimize token generation cycles more effectively, similar to out-of-order execution in CPU instruction scheduling. 

\subsection{Latency Analysis}
\label{ssec:math}
To understand how asynchronous function calling reduces latency, we conduct theoretical analyses under parallel function calling (\S\ref{ssec:parallel}). All proofs are available in \autoref{sec:proof}. 

\textbf{Overlapping generation and execution}. Asynchronous function calling is at least as fast as synchronous function calling.
The total latency for \sync in simple parallel function calling can be modeled as: $$L_{\sync}(F) = \textstyle\sum_{f \in F} G(f) + \sum_{f \in F} E(f),$$
given $F$ as a set of functions that do not depend on each other for execution, where $G(f)$ is the token generation latency for $f\in F$, and $E(f)$ is the execution time.
For \synccom, the total latency is:
$$L_{\synccom}(F) = \textstyle\sum_{f \in F} G(f) + \max_{f \in F} E(f),$$
assuming negligible overhead for parallelizing them.
For \async, using the LPT heuristic, the total latency is:
$$L_{\async}(F) = \max_{f \in F} \left( E(f) + \textstyle\sum_{g \in \mathrm{pred}(f,F)} G(g) \right),$$
where $\mathrm{pred}(f,F) = \{ g \in F \mid E(f) \leq E(g) \}$.
Intuitively, this formation indicates that each function call $f$ initiates after generating all function calls that has the same or higher priorities, i.e., sum of generation times in $\mathrm{pred}(f,F)$.
We prove that \async is at least as fast as \synccom in the worst case, assuming the number of generated tokens is the same and there are no additional overheads in \synccom or \async beyond $G(f)$s and $E(f)$s.
\begin{theorem}
\label{theorem:async_vs_synccom} For any set of independent functions $F$, 
$L_{\async}(F) \leq L_{\synccom}(F) < L_{\sync}(F)$.
\end{theorem}

\textbf{Characterizing speedups with average function execution times.} Assuming that the $E(\cdot)$ follows a normal distribution, we estimate the expected speedup of \async over \sync using $\overline{E}$ and $\overline{G}$, where $\overline{E}$ represents the average execution time, and $\overline{G}$ denotes the average generation time of functions in $F$.
\begin{theorem}
\label{theorem:estimation}
The ratio of speedup, $\frac{L_{\sync}}{L_{\async}}$ is approximately 
$ 1 + \frac{\overline{E}}{\overline{G}}$ with an error of $\mathcal{O}((\frac{\overline{E}}{\overline{G}})^2)$ when $|F|$ is large. 


\end{theorem}


This result indicates that tasks involving long average execution times (e.g., expensive I/Os) leads to a better speedup, whereas tasks with short execution times (e.g., simple arithmetic operations) may not benefit as much from \sys. 

\textbf{Rationale on using LPT heuristic.} LPT is optimal for parallel function calling.
We prove that the LPT heuristic minimizes total latency in asynchronous parallel function calling. Intuitively, starting the execution of longer functions earlier maximizes overlap with token generation, reducing the overall makespan. To put this argument formally,
\begin{theorem}
\label{theorem:lpt_optimality}
Continuing from Theorem~\ref{theorem:async_vs_synccom},
Any deviation from the LPT order cannot result in a lower total latency.
\end{theorem}
These analyses roughly demonstrate the advantage of asynchronous function calling, by overlapping generation and execution. as well as the rationale of LPT heuristic, especially when functions are independent of each other.

\subsection{System Overheads}\label{ssec:sys_overhead}

\textbf{Inference overhead of \sys.}
\sys introduces two potential sources of LLM inference overhead compared to \sync: (\textit{i}) syntactic overhead from using \dsl and function call identifiers, and (\textit{ii}) overhead from interrupts, caused by injecting function execution results and their identifiers into the LLM context.
The syntactic overhead is small since \sync and \synccom also require specific formats for function calls, e.g., wrapping function calls with markdown code syntax. \sys adds two or three control tokens per function call, which is negligible relative to the total tokens generated.
Similarly, the overhead from interrupts is mostly coming from in-context interrupt identifiers compared to \sync, since both methods receive function return values within the LLM context.
\async incurs an average of 20 additional tokens compared to \sync on the \code{v3-multi-step-parallel} dataset, adding about 90 ms of latency and 27.5 MB of additional KV cache in GPU memory on 3B model—less than 5\% of total memory usage.

\textbf{Performance without co-optimized the serving system.}
Implementing \sys on the OpenAI API (\asyncn) is impractical for GPT-4o due to high time-to-first-token (TTFT) latency. On GPT-4o, \asyncn suffers from 1.5\x higher latency than \synccom (\autoref{fig:multitask-latency}). The increased latency stems from the overhead of initiating a new API call for each interrupt, resulting in an average TTFT of 310 ms. Interestingly, \asyncn on the local Llama-3B does not experience such high TTFT, with an average TTFT of only 59 ms, making it 1.1\x faster than \synccom. Techniques like prompt caching or stateful inference may mitigate the TTFT overhead in \asyncn, potentially improving performance.

\begin{figure}[t]
    \centering
    \begin{subfigure}[b]{0.235\textwidth}
        \centering
        \includegraphics[width=\textwidth]{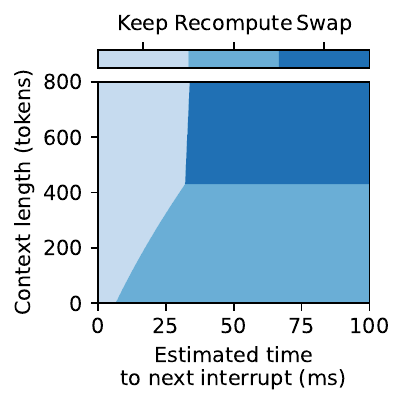}
        \caption{Llama-1B}
        \label{fig:trap_llama1B}
    \end{subfigure}
    \hfill
    \begin{subfigure}[b]{0.235\textwidth}
        \centering
        \includegraphics[width=\textwidth]{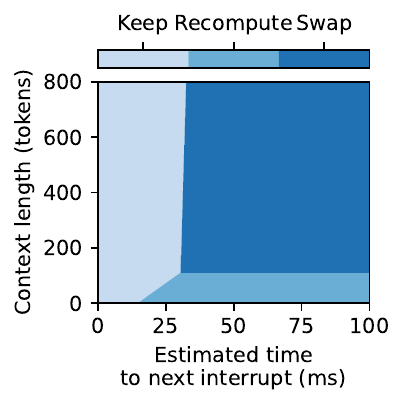}
        \caption{Llama-3B}
        \label{fig:trap_llama3B}
    \end{subfigure}
    \vspace{-0.8em}
    \caption{
        \textbf{The diagrams of the trap handling strategy} that minimizes idle KV cache in memory without sacrificing latency. The optimal strategy depends on the estimated time to the next interrupt and the number of tokens in the context. 
    }
    \label{fig:trap}
    \vspace{-1em}
\end{figure}

\textbf{LPT heuristic under task dependencies.}
Empirical experiments show that LPT is an effective heuristic. In our local \async evaluation in \autoref{fig:multitask-latency}, LPT was 8\% faster (on average) than randomly choosing the next function. 
However, we note that LPT is not guaranteed to be optimal when functions have dependencies. Consider functions $a$, $b$, and $c$, where $c$ depends on $a$, and their execution times satisfy $E(a) < E(b) < E(c)$. LPT would schedule them as $b$ -- $a$ -- $c$, since it does not consider future dependencies, while the optimal order is $a$ -- $c$ -- $b$. This challenge falls under resource-constrained project scheduling, an NP-hard problem~\cite{brucker1999resource,hartmann2022updated}.
LPT is a greedy heuristic that considers only currently available information without predicting future dependencies.

\textbf{Trap handling strategy.}
Trap handler can reduce idle KV cache memory usage by estimating the expected latency of swapping and recomputation based on the context length. \autoref{fig:trap} analyzes the optimal trap handling strategy for \sys on Llama-1B and Llama-3B models, using latency models of KV cache swap (linear) and recomputation (quadratic). For example, if the next interrupt is expected in 100 ms and the context has 300 tokens, the trap handler can drop the KV cache and recompute in 1B model, but should swap it out in 3B due to higher recomputation overhead.


\subsection{Function Calling Accuracy}
\label{ssec:accuracy}

To assess how \sys affects LLMs' ability to generate accurate function calls, i.e., correct signature, arguments, and calling order, we examine function call traces of each baseline using multi-step parallel dataset. We ensure all necessary function definitions are provided in the prompt. Function calling accuracy is measured by exact AST matching for each function call against the ground truth call, and their execution order.

We compare two LLM adaptation strategies: fine-tuning and few-shot prompting. Llama models are adapted using both methods, while GPT-4o is adapted with few-shot prompting only. To understand the effect of \dsl syntax, we also test a scenario where we format function calls in \sync in \dsl.
Table~\ref{table:accuracy} presents the results for GPT-4o and Llama models under these adaptation strategies. 

\begin{table}[t]
    \caption{Average function composition accuracy (AST matching) of \sync and \async in multi-step parallel function calling. (FT) denotes fine-tuned models; (ICL) denotes few-shot prompting.}
    \label{table:accuracy}
    \vspace{-1em}
    \begin{center}
    \begin{small}
    \begin{sc}
    \begin{tabular}{lccc}
    \toprule
    Model & Sync & Sync (in \dsl) & Async \\
    \midrule
    GPT-4o (ICL)             & 57.84\%  & 57.80\%  & 59.61\%  \\
    4o-mini (ICL)        & 55.08\%  & 53.41\%  & 41.44\%  \\
    \midrule
    Llama-3B (ICL)          & 16.98\%  & 17.33\%  & 5.46\%   \\
    Llama-1B (ICL)          & 0.49\%   & 0.39\%   & 1.45\%   \\
    \midrule
    Llama-3B (FT)      & 57.33\%  & 66.07\%  & 65.97\%  \\
    Llama-1B (FT)      & 6.33\%   & 12.15\%  & 12.06\%  \\
    \bottomrule
    \end{tabular}
    \end{sc}
    \end{small}
    \end{center}
    \vskip -0.1in
    \vspace{-1em}
\end{table}

\textbf{Impact of \dsl on accuracy.}
Formatting \sync responses using \dsl syntax has minimal effect on accuracy. In GPT-4o, accuracy remains virtually unchanged, while Llama-3B shows a slight improvement. Fine-tuned models exhibit a modest increase in accuracy when using \dsl, likely because the fine-tuning samples were formatted with \dsl.

\textbf{Impact of Asynchronous Function Calling on Accuracy.}
Both GPT-4o (with few-shot prompting) and fine-tuned Llama models maintain similar accuracy across \sync and \async, suggesting that asynchronous function calling does not degrade the models' function calling accuracy. However, Llama models adapted with few-shot prompting experience a significant drop in accuracy when switching from \sync to \async. We attribute this to the limited in-context learning capability of small models to adapt to \dsl syntax.


\textbf{Fine-tuning vs. Few-shot Prompting for LLM Adaptation.}
While GPT-4o adapts well to asynchronous function calling with few-shot prompting, smaller models like GPT-4o-mini experience a slight reduction in accuracy in \async. Few-shot prompted Llama models show low overall accuracy, whereas fine-tuning significantly improves their performance. Notably, fine-tuned Llama models achieve higher accuracy in \async compared to GPT-4o in \sync, highlighting the importance of fine-tuning for smaller LLMs.

%% file: sections/discussion.tex
\subsection{Discussion -- Human-triggered Interrupts}
\label{ssec:casestudy}

\begin{figure}
    \centering
    \includegraphics[width=0.9\columnwidth]{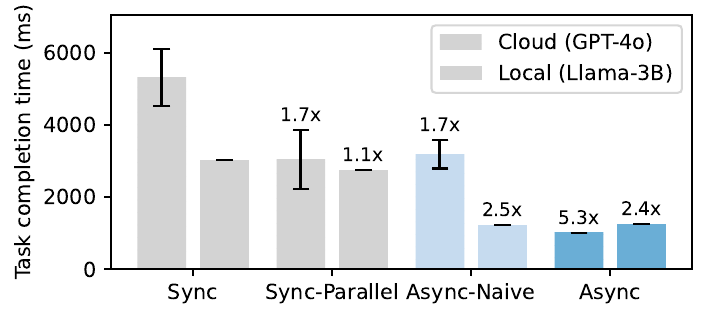}
    \vspace{-1em}
    \caption{\textbf{Evaluation on user-triggered interrupts}. This repeats the experiment in \autoref{fig:multitask-latency}, while each task is represented by an interrupt instead of a prompt, arriving at 0, 200, and 400 ms.}
    \label{fig:multi-turn-latency}
    \vspace{-1em}
\end{figure} 

The interrupt mechanism proposed in \sys not only allows asynchronous function calling, but also enables new types of human-LLM and LLM-LLM interactions.
We first discuss flexibility of this mechanism, then illustrate its potential with real-world examples.

\vspace{-5pt}
\paragraph{Generality and flexibility of interrupt.} 
\sys extends interrupt functionality beyond executor notifications, allowing interrupts to signal external events, such as user inputs or system triggers. This capability enables real-time interactions, where users can interrupt an ongoing LLM inference to add or adjust tasks without waiting for the current task to finish.
Although not detailed in this paper, we incorporated user-triggered interrupts into our fine-tuning dataset by randomly injecting new tasks from multi-turn conversation datasets. While this dataset may not perfectly represent real human-LLM interactions, we explored its effectiveness in the following scenarios.

\vspace{-5pt}
\paragraph{Interruptible AI assistants.}
LLM chatbots currently operate synchronously, requiring users to wait for complete responses. This approach increases perceived latency and limits real-time task handling. With \sys, users can issue new requests immediately, even during ongoing LLM responses. For instance, users often make follow-up requests without waiting for the initial response. A user might first ask to ``find a hotel in Seattle’’ and then quickly add ``near Space Needle.'' Similarly, a user can correct a previous request, such as saying ``Actually, make it Thursday’’ after initially scheduling a meeting for Wednesday.

To examine this scenario, we repeated the multi-step parallel function calling experiment, presenting tasks as a series of user-triggered interrupts rather than prompts. We inserted these interrupts every 200 ms and measured the end-to-end task completion latency of \async, as shown in Figure~\ref{fig:multi-turn-latency}. For comparison, we ran \sync and \synccom with the same tasks, each processing them sequentially. \async reduced latency by 2.4\x compared to \sync, whereas \synccom achieved only a 1.1\x reduction. The limited improvement of \synccom is because it requires generating the entire function calls at once, which is unsuitable for this scenario.

\paragraph{Multi-communicating LLM agents.} \sys also enables multiple autonomous LLM agents to communicate simultaneously. Typically, agent-to-agent communication uses synchronous message exchanges~\cite{wu2023autogen,chan2023chateval}, where agents take turns. This limits their ability to multitask and confines communication to a round-robin style. With \sys, one-to-many or many-to-many agent communication can be implemented using interrupt mechanisms. Each agent can send messages to others as interrupts, enabling more dynamic interactions, which could enable more realistic LLM-based social simulations~\cite{zhou2023sotopia,park2023generative}. This new communication pattern may introduce more complexity and require LLMs to learn when to chime in just like human conversation.

%% file: sections/conclusion.tex
\section{Conclusion}
\sys is a system that enables asynchronous LLM function calling by allowing the LLM and the function executor to operate independently. Our core innovation is making LLM inference interruptible through (1) \dsl, an in-context interface that facilitates asynchronous interaction, (2) adaptation of LLMs to leverage asynchronous semantics for optimized task completion latency, and (3) efficient implementation of an interrupt mechanism on the LLM inference pipeline. Empirical evaluations on the BFCL benchmark demonstrate that \sys reduces latency by 1.6x–5.4x compared to synchronous methods. \sys opens new possibilities for improving the operational efficiency of LLMs that interact with external tools, data sources, humans, and other LLMs.
\vspace{-0.5em}
\section*{Acknowledgments}
\noindent This work is supported in part by NSF Athena AI Institute (Award \#2112562) and Yale University.

%% file: sections/appendix.tex
\section*{Appendix}
\section{Latency Analysis}
\label{sec:proof}
In this section, we analyze the total latency of different execution models for a set of independent functions and provide proof sketches for the theorems that demonstrate the advantages of asynchronous function calling.

\renewcommand{\thetheorem}{6.\arabic{theorem}}

\subsection{Context and Definitions}

Let \( F = \{ f_1, f_2, \dotsc, f_n \} \) be a set of \( n \) independent functions. Each function \( f \in F \) has a token generation latency \( G(f) > 0 \) and an execution time \( E(f) > 0 \). We define the average generation latency \( \overline{G} = \frac{1}{n} \sum_{f \in F} G(f) \) and the average execution time \( \overline{E} = \frac{1}{n} \sum_{f \in F} E(f) \).

The total latencies under different execution models are as follows.

\emph{ $\bullet$ Synchronous execution (\( \sync \))}: Tokens are generated and functions are executed sequentially:
\[ L_{\sync}(F) = \sum_{f \in F} G(f) + \sum_{f \in F} E(f) = n (\overline{G} + \overline{E}). \]

\emph{ $\bullet$ Synchronous with concurrent execution (\( \synccom \))}: Tokens are generated sequentially, functions are executed concurrently:
\[ L_{\synccom}(F) = \sum_{f \in F} G(f) + \max_{f \in F} E(f). \]

\emph{ $\bullet$ Asynchronous execution (\( \async \))} using the \textit{Longest Processing Time (LPT)} heuristic: Tokens are generated sequentially, functions are scheduled in decreasing order of \( E(f) \):
\[ L_{\async}(F) = \max_{f \in F} \left( E(f) + \sum_{g \in \mathrm{pred}(f,F)} G(g) \right), \]
where \( \mathrm{pred}(f,F) = \{ g \in F \mid E(f) \leq E(g) \}. \)

\subsection{Proof Sketch of Theorem 6.1}

\begin{theorem}
For any set of independent functions \( F \),
\[ L_{\async}(F) \leq L_{\synccom}(F) < L_{\sync}(F). \]
\end{theorem}

1. Establishing \( L_{\synccom}(F) < L_{\sync}(F) \):

Since the maximum execution time is less than the sum of all execution times,
\[ \max_{f \in F} E(f) < \sum_{f \in F} E(f), \]
it follows that
\[ L_{\synccom}(F) < \sum_{f \in F} G(f) + \sum_{f \in F} E(f) = L_{\sync}(F). \]

2. Establishing \( L_{\async}(F) \leq L_{\synccom}(F) \):

Under the LPT heuristic, functions are ordered such that \( E(f_1) \geq E(f_2) \geq \dotsb \geq E(f_n) \). The completion time of function \( f_i \) is
\[ C(f_i) = \left( \sum_{j=1}^{i} G(f_j) \right) + E(f_i). \]
The total latency is \( L_{\async}(F) = \max_{1 \leq i \leq n} C(f_i) \). Since for all \( i \),
\[ \sum_{j=1}^{i} G(f_j) \leq \sum_{f \in F} G(f), \quad E(f_i) \leq E(f_1) = \max_{f \in F} E(f), \]
we have
\[ C(f_i) \leq \sum_{f \in F} G(f) + \max_{f \in F} E(f) = L_{\synccom}(F), \]
and thus \( L_{\async}(F) \leq L_{\synccom}(F). \)

\subsection{Proof Sketch of Theorem 6.2}

\begin{theorem}
The ratio \( \frac{L_{\sync}(F)}{L_{\async}(F)} \) is approximately \( 1 + \frac{\overline{E}}{\overline{G}} \) with an error order of \( \left( \frac{\overline{E}}{\overline{G}} \right)^2 \) when \( n = |F| \) is large.
\end{theorem}

When \( n \) is large, we can approximate the cumulative token generation latency and the maximum execution time using order statistics, assuming \( E(f) \) follows a normal distribution:
\[ \sum_{f \in F} G(f) \approx n \overline{G}, \quad \max_{f \in F} E(f) \approx \overline{E} + \sigma \sqrt{2 \ln n}. \]
Using the inequality \( L_{\async}(F) \leq L_{\synccom}(F) \), we approximate the total latency as
\[ L_{\async}(F) \approx n \overline{G} + \overline{E} + \sigma \sqrt{2 \ln n}. \]

The ratio of total latencies is then
\[ \frac{L_{\sync}(F)}{L_{\async}(F)} \approx \frac{n \overline{G} + n \overline{E}}{n \overline{G} + \overline{E} + \sigma \sqrt{2 \ln n}}. \]
Since \( \varepsilon = \dfrac{\overline{E} + \sigma \sqrt{2 \ln n}}{n \overline{G}} \ll 1 \) for large \( n \), we can approximate
\( \dfrac{1}{1 + \varepsilon} \approx 1 - \varepsilon \). Applying this approximation,
\[ \frac{L_{\sync}(F)}{L_{\async}(F)} \approx \left( 1 + \dfrac{\overline{E}}{\overline{G}} \right) \left( 1 - \dfrac{\overline{E} + \sigma \sqrt{2 \ln n}}{n \overline{G}} \right). \]
Neglecting terms of order \( O(1/n) \), we obtain
\[ \frac{L_{\sync}(F)}{L_{\async}(F)} \approx 1 + \dfrac{\overline{E}}{\overline{G}}. \]

The error in this approximation is of order \( \left( \dfrac{\overline{E}}{\overline{G}} \right)^2 \).


\subsection{Proof Sketch of Theorem 6.3}

\begin{theorem}
Any deviation from the LPT order cannot result in a lower total latency.
\end{theorem}

Assume, for contradiction, that there exists a schedule \( \sigma' \) that deviates from the LPT order and yields a lower total latency than the LPT schedule \( \sigma^* \). This implies that in \( \sigma' \), there exist functions \( f_i \) and \( f_j \) such that \( E(f_i) < E(f_j) \) but \( f_j \) is scheduled after \( f_i \).

Let the cumulative token generation latency before \( f_i \) and \( f_j \) in \( \sigma' \) be \( S_i \) and \( S_j \), respectively, with \( S_j = S_i + G(f_i) \). The completion time is:
\[ C_{\sigma'}(f_j) = S_j + G(f_j) + E(f_j) = S_i + G(f_i) + G(f_j) + E(f_j). \]

Consider swapping \( f_i \) and \( f_j \) to obtain schedule \( \sigma'' \). The sum of token generation latencies become \( S_j'' = S_i \) for \( f_j \), and \( S_i'' = S_j'' + G(f_j) \) for \( f_i \). The new completion time is
\[ C_{\sigma''}(f_i) = S_i'' + G(f_i) + E(f_i) = S_i + G(f_j) + G(f_i) + E(f_i). \]

Comparing the completion times before and after the swap:
\[ C_{\sigma''}(f_i) - C_{\sigma'}(f_i) = E(f_i) - E(f_j) < 0. \]
Thus, swapping \( f_i \) and \( f_j \) always reduces the maximum completion time. By repeatedly applying such swaps, any schedule can be transformed into the LPT schedule without increasing the total latency. This contradicts the assumption that \( \sigma' \) yields a lower total latency than \( \sigma^* \). Therefore, any deviation from the LPT order cannot result in a lower total latency.